\documentclass[sigconf, nonacm]{acmart}

\usepackage{multirow}
\usepackage{amsmath}

\newcommand{\tagthink}{\texttt{<think>}}
\newcommand{\tagthinkend}{\texttt{</think>}}
\newcommand{\taganswer}{\texttt{<answer>}}
\newcommand{\taganswerend}{\texttt{</answer>}}
\newcommand{\tagclass}{\texttt{<class>}}
\newcommand{\tagclassend}{\texttt{</class>}}
\newcommand{\tagtimestep}{\texttt{<timestep>}}
\newcommand{\tagtimestepend}{\texttt{</timestep>}}

\AtBeginDocument{%
  }

\begin{document}

\title[Complementing Text Moderation with Grounding]{Complementing Text Based Moderation with Multimodal Grounding: Detecting Fraudulent Inducement Behaviors on Short Video Platforms}

\author{Xiaoyong Yu}
\affiliation{%
  \institution{Huazhong University of Science and Technology}
  \city{Wuhan}
  \country{China}
}
\email{xiaoyongyu666@gmail.com}

\renewcommand{\shortauthors}{X. Yu}

\begin{abstract}
Short video platforms have become a primary channel for black and gray industry operations in internet finance, where inducement behaviors such as prompts to like, follow, join private chat groups, or click external links frequently steer users toward fraud. Because such videos are short, dynamic, and often convey intent through fleeting visual cues rather than stable text, a machine assisted moderation system for this domain must guarantee detection recall so that no inducement is silently missed, and must localize the offending segment so that reviewers inspect only the relevant clip. Existing moderation system is an Optical Character Recognition (OCR) and Large Language Model (LLM) pipeline that transcribes on screen text frame by frame and classifies the transcript. It is robust and deployed at scale, but it can only exploit textual information: many inducements are conveyed through visual cues that OCR cannot observe, and it cannot localize when an inducement occurs beyond the keyframes where its text appears. We explore \textbf{OCR+MLLM}, which augments the text based pipeline with a Multimodal Large Language Model (MLLM), and systematically compare three routes: the internal OCR+LLM system (keyframe based localization), OCR+MLLM (learned temporal grounding), and a large zero shot MLLM, along detection, latency, visual token consumption, and deployment complexity. OCR preserves the text signal that guarantees recall, while the MLLM jointly performs OCR grounded discrimination and temporal video grounding, localizing both textual and non textual inducements. To train the MLLM under severe class imbalance and sparse temporal supervision, we construct a video grounding dataset of internet finance short videos spanning internet credit, investment, and insurance, comprising 2,419 videos (66.49 hours) and 3,194 finely annotated events across four categories, and systematically study reward design for reinforcement learning (RL) based grounding with classification. We report two transferable findings: auxiliary reward terms are systematically exploited through reward hacking, and, under class imbalance, stronger localization rewards degrade class balanced detection. We therefore design a recall weighted (F$_2$) localization reward with class balanced resampling. Experiments show that OCR+MLLM localizes clearly better than the keyframe based OCR+LLM baseline (mean Intersection over Union (mIoU) 26.52 vs. 14.49) and detects and localizes non textual inducements that text only pipelines miss, while maintaining 67.0\% macro recall.
\end{abstract}

\maketitle



\section{Introduction}
As short video platforms have become a primary channel for black and gray industry operations in internet finance, inducement behaviors, such as prompting users to like, follow, join private chat groups, or click external links, frequently steer users toward fraud~\cite{lang2025biting,wang2025crossmodal,yu2026whenrules,lu2025vlm,9552250,liu-etal-2026-ips}. These behaviors are conveyed through subtle cues: a fleeting hand gesture, a small or missing on screen caption, a brief flash of a QR code, or a suggestive background audio announcement. Because such cues are implicit and context dependent, they are difficult for both human moderators and automated systems: human review does not scale to platform volumes~\cite{lu2025vlm}, and automated systems lack the temporal precision to isolate the offending segment. Recent work shows that multimodal models can detect inappropriate short video content at scale~\cite{wang2025reasoning}, that video based cues can predict financial fraud~\cite{duan2024video}, and that multimodal, policy aware detection can surface scam videos beyond metadata~\cite{kulsum2026beyond}; financial fraud detection itself has been surveyed across architectures and deployment challenges~\cite{thivaios2026fraudsurvey}; yet covert financial inducements remain largely unaddressed. For internet finance in particular, inducements are embedded in loan, fund, insurance, and investment content, where a single missed case can cause direct financial loss; detecting them together with their temporal locations is therefore the central requirement of a machine assisted moderation architecture for this domain.

\begin{figure*}[htbp]
    \centering
    \includegraphics[width=\linewidth]{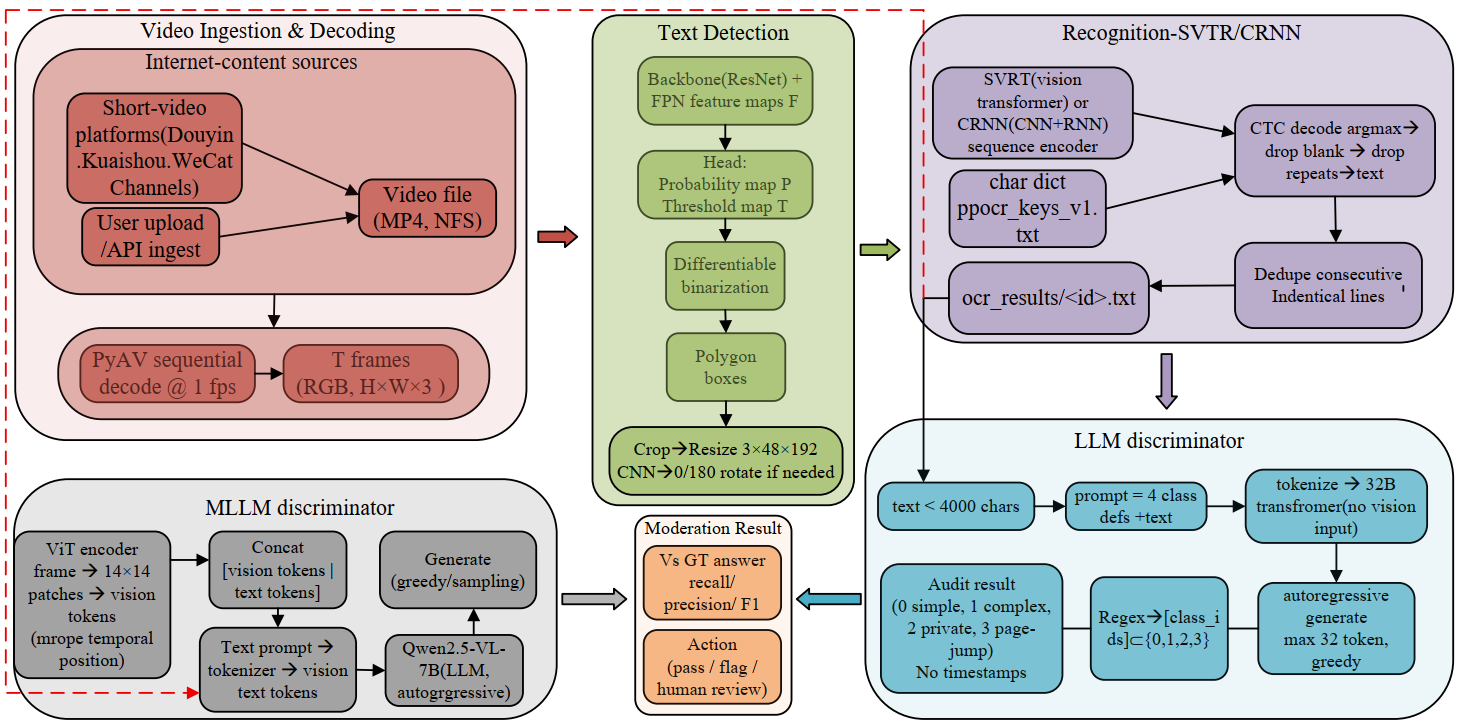}
    \caption{Comparison of the OCR+LLM moderation pipeline and the OCR+MLLM moderation framework.}
    \label{fig:cot_pipeline}
\end{figure*}

Because internet finance is a highly specialized domain, its short videos contain substantial on screen text, and existing moderation systems generally adopt an OCR+LLM architecture. We take as our starting point such an internal system, a two stage OCR and LLM pipeline (Figure~\ref{fig:cot_pipeline}): on screen text is transcribed frame by frame and the transcript is classified. While robust and deployed at scale, it has a structural blind spot: it perceives only inducements \emph{expressed through text}. Many inducements, such as a host gesturing toward a like button or a brief QR code flash, leave no textual trace that OCR can observe. It also localizes inducements only by the keyframes where their text appears, so it misses non textual inducements entirely and cannot recover precise temporal boundaries; recent surveys confirm that precise localization remains an open challenge for MLLMs~\cite{wu2026survey}. We stress that our goal is not to discard OCR, which reliably reads small, transient, and rapidly changing on screen text and keeps recall high. What OCR lacks is the ability to see visual cues and to place an event in time. Our contribution is therefore an improvement of the OCR+LLM system: we retain OCR as the text channel and add a multimodal model that complements it on the two dimensions it cannot cover.

Prior work has explored multimodal approaches for safety critical short video tasks, including toxic content detection via text and audio features~\cite{supriyanto2026automatic}, federated multimodal learning for imbalanced fake news detection~\cite{guofed}, cross modal consistency modeling for fake news~\cite{tian2026exposing}, and intent aware modeling that mitigates shortcut learning~\cite{kong2026said}. Multimodal frameworks have also been adopted for large scale video understanding~\cite{zha2026interpreting,miehling2026multimodal,paekivi2026modeling,rodas2026quality}, and MLLMs have become powerful tools for short video analysis~\cite{gorniscalable,liu2026interpreting,soltan2026automated,liu2026bridging,meng2026watch,zhang2026dynframe}. Temporal video grounding has advanced through dataset construction and paradigm design~\cite{zheng2026omnivtg,li2026universal,zhang2026timelens,jin2026should}, including reasoning guided~\cite{wang2025timezero,wang2026time}, spatio temporal~\cite{wang2026spacevllm,yang2026unleashing}, and RL based grounding~\cite{bai2025univg,wang2025video}. Yet existing methods remain inadequate for covert inducements in short videos~\cite{yang-etal-2026-univid}: distinguishing normal showcases from covert inducements requires jointly comprehending subtle visual cues, suggestive audio, and implicit textual signals, a challenge that single modality or shallow fusion approaches cannot handle, and the field lacks high quality annotated datasets.

In this paper, we augment our internal OCR+LLM system with temporal grounding and compare three routes, the internal OCR+LLM system (keyframe based localization), OCR+MLLM (learned temporal grounding), and a large zero shot MLLM, along detection performance, latency, visual token consumption, and deployment complexity. OCR+MLLM consists of two complementary components: (i) an OCR stage that extracts on screen text, preserving the text signal that underlies reliable recall; and (ii) a 7B MLLM that takes both the video and the OCR transcript and jointly predicts each inducement event's category and approximate temporal region in a single forward pass. The MLLM thus plays a dual role: OCR grounded discrimination, inheriting the recall of the text based pipeline, and temporal video grounding, localizing inducements conveyed purely visually. All compared methods share the same OCR stage; only our framework trains the downstream model to perform grounding. OCR handles textually expressed inducements, and the MLLM handles non textual inducements and temporal localization, supplementing OCR exactly where OCR is blind. Critically, we treat grounding not as an end in itself but as a means to reduce human review workload under a hard safety constraint: detection recall must be at least as good as the text only baseline before any gain in grounding precision is credited. We construct a video grounding dataset covering four inducement categories with fine grained temporal boundaries across multiple financial sub domains, and systematically study reward design for RL based grounding with classification under severe class imbalance and sparse temporal supervision, culminating in a recall weighted reward and class balanced resampling that realize the recall first objective.

Our study reveals two findings generalizable beyond this task. First, auxiliary reward terms, center distance penalties and event count rewards, are exploited by reward hacking: the model maximizes the auxiliary signal without improving true localization, inflating training reward while degrading real Intersection over Union (IoU). Second, a fundamental tradeoff exists between localization and class balanced detection under class imbalance: multiplicative coupling between IoU and class rewards learns localization but collapses predictions to the dominant class, whereas additive coupling improves class balance but stalls localization. We trace this tradeoff to the tension between IoU sparsity (no gradient when intervals do not overlap) and class imbalance (the simple interaction class accounts for roughly 62\% of events).

The main contributions are:
\begin{itemize}
    \item We introduce the task of detecting and localizing fraudulent inducements in internet finance short videos, a covert, high stakes moderation problem that existing systems address only through text, and construct a fine grained benchmark of 2,419 videos with 3,194 annotated events across four categories.
    \item We propose \textbf{OCR+MLLM}, which augments OCR with a 7B MLLM that jointly performs OCR grounded discrimination and temporal grounding, and identify the two failure modes it addresses: OCR misses non textual inducements, and a grounding MLLM alone degrades on mostly static frames.
    \item We systematically study reward design for RL based grounding with classification, revealing auxiliary term reward hacking and a localization class balance tradeoff, and identify a class scaling sweet spot that partially mitigates it.
    \item Comparing OCR+LLM, OCR+MLLM, and a large zero shot MLLM along detection, latency, token consumption, and deployment complexity, we show that OCR+MLLM matches OCR+LLM on classification and localization while adding the decisive ability to detect and localize \emph{non textual} inducements, reducing reviewer workload without compromising detection safety.
\end{itemize}

\section{Related Work}
\subsection{Content Moderation on Web Platforms}
Short video moderation research integrates text and audio features for toxic content detection~\cite{supriyanto2026automatic}. In video moderation, In Prompt Process Supervision (IPS)~\cite{liu-etal-2026-ips} improves short video moderation, the Unified Vision Language Model for Video Moderation (UNIVID)~\cite{yang-etal-2026-univid} jointly handles multiple moderation tasks, retrieval augmented multimodal experts target short video hate detection~\cite{lang2025biting}, cross modal transfer from memes to videos addresses hateful video detection~\cite{wang2025crossmodal}, and agent driven discovery surfaces emerging content issues~\cite{yu2026whenrules}. Federated multimodal learning~\cite{guofed} addresses cross platform heterogeneity and class imbalance in fake news detection, MAGIC3~\cite{tian2026exposing} models cross modal inconsistencies for interpretable fake news detection, intent aware approaches~\cite{kong2026said} mitigate shortcut learning, and cascade designs~\cite{wang2025filter} make MLLM based moderation practical at industrial scale. Multimodal sensor fusion has been applied to fraud perception in short video and live streaming platforms~\cite{zhao2026multimodal}. None addresses fraudulent inducements in internet finance short videos, which are conveyed through subtle gestures, sparse on screen text, brief QR code flashes, and suggestive audio, and require precise temporal localization for clip level evidence. Our OCR+MLLM framework fills this gap by augmenting the OCR+LLM pipeline with a 7B MLLM post trained via RL for joint classification and temporal grounding.

\subsection{Multimodal Large Language Models}
MLLMs have been employed for scalable content analysis via video to text pipelines~\cite{gorniscalable}, for interpreting social media videos~\cite{liu2026interpreting}, for automated creative evaluation of advertising content~\cite{soltan2026automated}, and for short drama quality assessment through a two stage framework combining Supervised Fine Tuning (SFT) with Group Relative Policy Optimization (GRPO) based preference alignment~\cite{liu2026bridging}. Existing methods remain inadequate for covert inducements in short videos.

\subsection{Reinforcement Learning Post Training for MLLMs} GRPO~\cite{shao2024deepseekmath} replaces a learned critic with group normalized advantages, reducing training cost relative to Proximal Policy Optimization (PPO) and enabling verifiable reward RL at scale, and has been widely adopted for reasoning oriented post training~\cite{zhou2026reinforced}. In the video domain, TimeZero~\cite{wang2025timezero} employs a think then answer paradigm via GRPO, Time-R1~\cite{wang2026time} introduces time aware reinforcement fine tuning, UniVG R1~\cite{bai2025univg} applies reasoning guided RL to universal visual grounding, and Video Thinker~\cite{wang2025video} generates key clues during inference; GRPO based preference alignment has also been used for short drama quality assessment~\cite{liu2026bridging}. Reward design remains underexplored, especially the interaction among multiple reward objectives under class imbalance and sparse temporal supervision~\cite{zhou2026reinforced,rlmultimodal2026survey,zeng2026rewardhacking,wang2026incentivizing}, which we study systematically for joint localization and classification.

\subsection{Temporal Video Grounding}
Temporal video grounding has advanced through dataset construction and paradigm design: OmniVTG~\cite{zheng2026omnivtg} proposes a large scale open world dataset with self correcting training, UniTime~\cite{li2026universal} enables universal grounding via adaptive frame scaling and coarse to fine localization, and TimeLens~\cite{zhang2026timelens} rethinks Video Temporal Grounding (VTG) from data quality and algorithm design. Output paradigm analysis~\cite{jin2026should} shows that continuous distribution based decoding achieves the best efficiency accuracy tradeoff, universal grounding~\cite{li2026universal} extends this to adaptive frame scaling, while for spatio temporal grounding, SpaceVLLM~\cite{wang2026spacevllm} uses a decoupled head architecture, a zero shot framework~\cite{yang2026unleashing} leverages decoupled spatio temporal highlighting, and recent surveys~\cite{wu2026survey} map the remaining open problems.

\section{Short Video Inducement Behavior Recognition Dataset}
\subsection{Construction and Annotation}
To reflect real world challenges in financial content governance, we curated a large scale video collection from two dominant short video platforms, Douyin and Kuaishou. We screened approximately 20,000 internet finance videos, from which we identified 2,419 videos exhibiting inducement behaviors as the positive set. The data spans internet credit, internet funds, real estate investment, internet insurance, knowledge popularization, financial news, and anti fraud publicity. Loan intermediaries, post loan consultation, loan advertisements, investment intermediaries, and insurance intermediaries exhibit the highest inducement likelihood, whereas pure knowledge popularization is not treated as inducement unless explicit cues such as credit report terminology appear. To mitigate catastrophic forgetting and reflect realistic business traffic, a small proportion of ordinary lifestyle videos is included in the negative set. We categorize inducements into four fine grained classes (Table~\ref{tab:taxonomy}) forming a progression of increasing moderation risk: Class 0 (simple interaction) is largely benign engagement farming; Class 1 (complex interaction) is intermediate; Class 2 (private conversion) solicits off platform contact through QR codes or group IDs and is the most fraud sensitive; and Class 3 (page transition) covers link clicks and app installs. Orthogonal to this taxonomy, an inducement can be conveyed either \emph{textually}, through on screen text such as ``follow'' or ``scan the QR code'', or \emph{non textually}, through purely visual cues such as a QR code flash or a gesture toward a like button, leaving no textual trace. A text only pipeline observes only the former, whereas a multimodal model observes both. We determine each event's modality by comparing its class label against the per frame OCR transcript of its temporal window. At the video level, a video is \emph{non textual} if at least one event is non textual, and \emph{textual} only when every event is textual.

\begin{table}[htbp]
    \centering
    \caption{Fine grained inducement behavior taxonomy.}
    \label{tab:taxonomy}
    \small
    \begin{tabular}{clp{5.0cm}}
        \toprule
        \textbf{ID} & \textbf{Class} & \textbf{Description} \\
        \midrule
        0 & Simple interaction & Likes, favorites, shares, follows, @ mentions, screenshot saving \\
        1 & Complex interaction & Comments, DMs, live stream interaction, one on one consultation \\
        2 & Private conversion & WeChat/QQ group ID or QR code, fan group joins, memberships \\
        3 & Page transition & Clicking bottom links, installing apps, searching software \\
        \bottomrule
    \end{tabular}
\end{table}

\begin{figure*}[htbp]
    \centering
    \includegraphics[width=\linewidth]{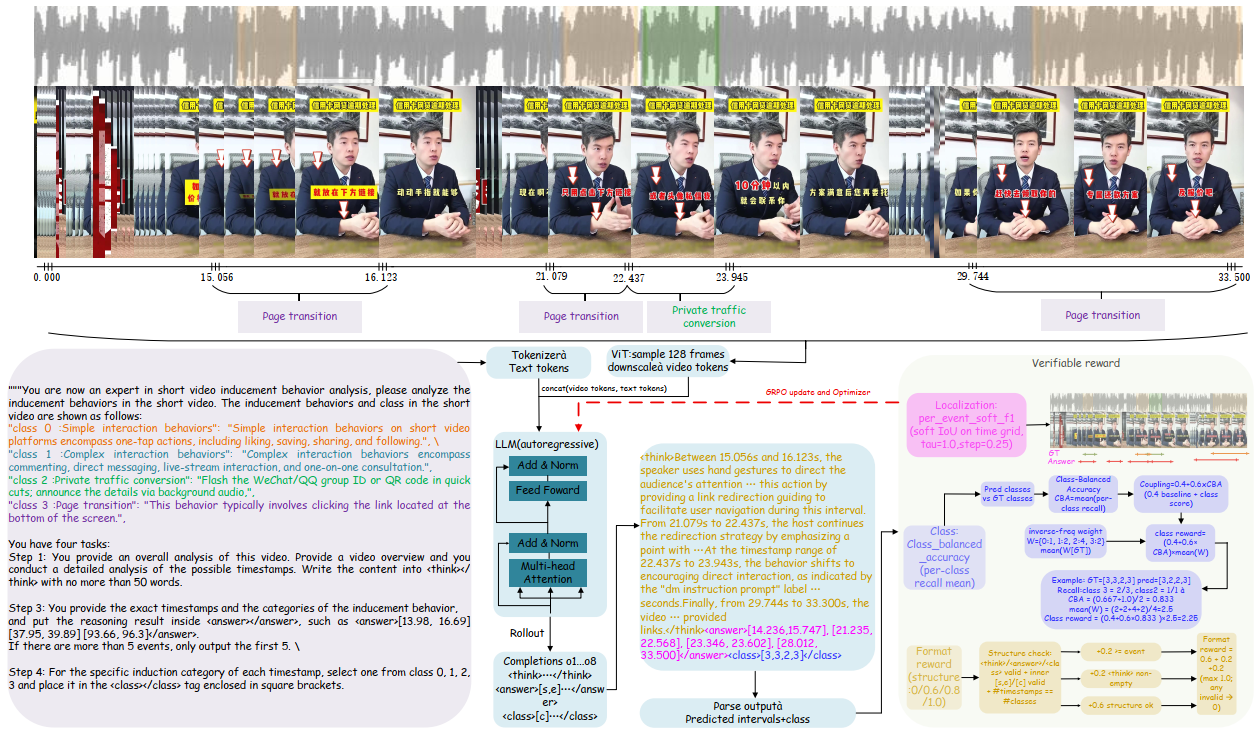}
    \caption{Overview of the RL post training pipeline for the grounding model.}
    \label{fig:vlm-train}
\end{figure*}

\subsection{Dataset Statistics}
The dataset contains 2,419 videos (train: 1,694; test: 725), totaling 66.49 hours with 3,194 annotated events. Each sample includes the video path, duration, timestamp intervals, class labels, and optional clipping boundaries. Table~\ref{tab:class-dist} reports class distribution at event and video levels; video level counts are not mutually exclusive, as a video may contain events from multiple classes. Simple interaction dominates with 62.3\% of test events, followed by complex interaction (22.7\%) and page transition (11.5\%), while private conversion is rarest at 3.6\%; this long tailed distribution motivates the class balanced reward design in Section~\ref{sec:reward}.

Using the per frame OCR transcript, we label each event as \emph{textual} (its window contains class indicative keywords such as ``follow'' or ``scan the QR code'') or \emph{non textual} (no such keywords; conveyed purely visually). On the test set, approximately 50.4\% of events are non textual, spanning 458 of 725 videos (63\%). The non textual share is similar across classes (37.2\%--53.8\%), so the coverage gap is not confined to a single category.

Table~\ref{tab:duration} summarizes event density and duration. Videos contain 1.30 and 1.36 events on average in the train and test splits, with a maximum of 6 events in a single test video. Test videos range from 11.1 to 532.0 seconds (mean 106.3 s), indicating substantial variation in the temporal scale at which inducements must be localized.

\begin{table*}[htbp]
    \centering
    \caption{Class distribution of the training and test sets (event level and video level).}
    \label{tab:class-dist}
    \small
    \begin{tabular}{lcccc}
        \toprule
        \textbf{Class} & \textbf{Train \#Events} & \textbf{Train \#Videos} & \textbf{Test \#Events} & \textbf{Test \#Videos} \\
        \midrule
        0 Simple interaction & 1,258 & 1,136 & 613 & 527 \\
        1 Complex interaction & 435 & 414 & 223 & 208 \\
        2 Private conversion & 135 & 61 & 35 & 31 \\
        3 Page transition & 382 & 266 & 113 & 84 \\
        \bottomrule
    \end{tabular}
\end{table*}

\begin{table}[htbp]
    \centering
    \caption{Event density and duration statistics.}
    \label{tab:duration}
    \small
    \begin{tabular}{lccc}
        \toprule
        \textbf{Stat.} & \textbf{Train Ev.} & \textbf{Test Ev.} & \textbf{Test Dur. (s)} \\
        \midrule
        Min & 1 & 1 & 11.1 \\
        Max & 14 & 6 & 532.0 \\
        Mean & 1.30 & 1.36 & 106.3 \\
        \bottomrule
    \end{tabular}
\end{table}

\section{Training Process and Optimization Strategy}
\subsection{Framework Overview}
Figure~\ref{fig:vlm-train} gives an overview of the proposed OCR+MLLM framework. Given a short video, the framework first runs an OCR stage (PaddleOCR PP OCRv4, where PP denotes PaddlePaddle) to transcribe on screen text frame by frame into a compact transcript. The transcript is concatenated with the video and fed to a 7B MLLM, which jointly (i) discriminates inducement categories from the OCR transcript and visual content, and (ii) grounds each event to its temporal interval. This design is an augmentation of the existing OCR+LLM pipeline rather than a departure from it: OCR is retained unchanged as the text channel, and the MLLM is added to supply visual understanding and temporal grounding. In contrast, OCR+LLM can only exploit textual information, aggregating the transcript into a class label and discarding the visual scene and temporal structure, whereas the MLLM integrates text, vision, and time into a single grounded prediction. Formally, the MLLM produces a structured output of a reasoning trace, timestamp intervals, and class labels:
\begin{equation*}
\begin{gathered}
\tagthink\ \cdots\ \tagthinkend \\
\taganswer\ [s_1,e_1]\cdots[s_k,e_k]\ \taganswerend \\
\tagclass\ [c_1,\ldots,c_k]\ \tagclassend
\end{gathered}
\end{equation*}

The structured output is scored by four reward terms, whose design we detail next.

\subsection{Reward Design}\label{sec:rewarddesign}
Reward design is the core of our post training recipe, because the task couples two objectives that pull in different directions: temporal grounding, which requires tight interval alignment, and classification, which requires correct category assignment under severe class imbalance. We define four terms, each targeting one failure mode we observed in preliminary runs.

Prior work shows that hybrid multi-aspect rewards outperform monolithic scalar rewards~\cite{gulhane2026hybrid}. The four terms are not independent: the format reward makes the output scorable at all, the IoU reward supplies the only dense signal for interval boundaries, the class balanced reward counteracts the long tail, and the process reward shapes the reasoning trace that the other three read. Removing any one of them collapses a different objective, which is why we treat them as a set rather than tuning a single scalar reward.

\textbf{Format reward.} The format reward enforces structural validity through three hard gates: the output must match the outer tag pattern, every interval inside \taganswer{} must be a well formed pair $[a, b]$ with $a < b$, and the number of intervals must equal the number of class labels. A valid output yields
\begin{equation}
R_{fmt} = 0.6 + 0.2 \cdot \mathbb{I}[\text{think} \neq \emptyset] + 0.2 \cdot \mathbb{I}[k > 0],
\end{equation}
where $k$ is the number of predicted events, while any gate violation yields $R_{fmt} = 0$. Without this term, the model drifts toward free form descriptions that are hard to parse and impossible to score.

\textbf{Multi interval IoU reward.} The grounding term merges predicted and ground truth intervals to eliminate overlaps and computes
\begin{equation}
R_{IoU} = \frac{\text{Inter}}{\text{Union}},
\end{equation}
where $\text{Inter}$ and $\text{Union}$ are the total intersection and union duration of the merged sets, with $R_{IoU} = 1$ when both sets are empty and $R_{IoU} = 0$ when exactly one is empty. The merge step is what allows the reward to support videos with up to 14 events without penalizing a single merged region multiple times.

\textbf{Class balanced accuracy reward.} Because Class 0 accounts for roughly 62\% of events while Class 2 accounts for 3.6\%, a plain accuracy term collapses to the dominant class. We therefore compute per class recall and average uniformly over classes present in the ground truth,
\begin{equation}
R_{cls} = \frac{1}{|\mathcal{C}|} \sum_{c \in \mathcal{C}} \frac{\text{TP}_c}{\text{TP}_c + \text{FN}_c},
\end{equation}
where $\mathcal{C}$ is the set of classes present and $\text{TP}_c$, $\text{FN}_c$ are matched and missed events under greedy one to one assignment. This assigns equal weight to each present class regardless of frequency and is the term that rescues the rarest class from collapse.

\textbf{Process supervision reward.} The process term scans the chain of thought within \tagthink{} tags for \tagtimestep{} start to end \tagtimestepend{} patterns and caps their count,
\begin{equation}
R_{process} = w \cdot \min(N_{\text{pairs}}, N_{\text{max}}),
\end{equation}
where $w=0.2$ and $N_{\text{max}}=1$. The cap matters: without it, the model emits redundant timestep annotations to accumulate reward, which is the auxiliary term reward hacking we analyze in Section~\ref{sec:reward}. Together with the class balanced term, the cap keeps the reward focused on genuine localization rather than on annotation volume, a distinction we quantify in Section~\ref{sec:reward}.

\subsection{Reward Coupling and Curriculum Learning}
The final reward combines these terms through multiplicative or additive coupling:
\begin{align}
R_{\text{mul}} &= R_{IoU} \cdot \left(a + b \cdot R_{cls}\right) \\
R_{\text{add}} &= R_{IoU} + R_{cls}
\end{align}
where $a,b\ge 0$ with $a+b=1$ (we use $a=0.3$, $b=0.7$). Our ablation (Section~\ref{sec:reward}) shows that multiplicative coupling learns localization but collapses to Class 0, while additive coupling improves class balance but stalls localization, a tradeoff rooted in the tension between IoU sparsity and class imbalance, consistent with the multi-objective reward hacking that MO-GRPO identifies and mitigates~\cite{ichihara2026mogpo}.

To stabilize training, we implement conditional data shuffling governed by a curriculum learning flag; when enabled, the pre sorted difficulty progression is preserved. We freeze the pre trained Vision Transformer (ViT) to reduce memory consumption while preserving the trainability of the visual merger layer, integrate Low Rank Adaptation (LoRA), and configure strict resolution thresholds to prevent out of memory errors.

\section{Experimental Evaluation}
\subsection{Experimental Setup}\label{sec:setup}
\textbf{Models.} We compare three routes. The \emph{OCR+LLM} route is the production two stage pipeline: PaddleOCR PP OCRv4 (DBNet + SVTR, CPU, FP32) at 1fps, followed by a text only LLM (Qwen2.5 VL 7B Instruct) that classifies the transcript (BF16, greedy decoding), localizing inducements by the keyframes where their text appears. The \emph{OCR+MLLM} route (ours) augments the same OCR stage with a grounding MLLM, Qwen2.5-VL-7B-Instruct, which takes frames and the OCR transcript and is trained with BF16 and FlashAttention 2 to jointly discriminate classes and localize events. The \emph{Large MLLM} route is a Qwen2.5-VL-32B-Instruct used zero-shot as a scale reference. As an ablation, we train the grounding MLLM \emph{without} the OCR transcript (MLLM only) to isolate OCR's contribution to recall.

\textbf{Data.} Training uses 1,694 videos with 2,210 events (45.08 h); testing uses 725 videos with 984 events (21.41 h). After removing undecodable videos (train 79, test 42), 2,298 videos and 3,045 events remain. Unless noted, we report on the full 725 video test set; the grounding route covers the 683 decodable videos, while OCR+LLM covers all 725.

\textbf{RL algorithm and hyperparameters.} We use GRPO with group size 8, KL coefficient 0.0 (disabled), clipping 0.2, learning rate $1\times10^{-6}$ with linear scheduling and no warmup, AdamW ($\beta=(0.9,0.999)$, weight decay 0.0), max gradient norm 1.0, and 2 epochs with checkpoints every 50 steps. The effective global batch size is 8 (per device batch 1, gradient accumulation 2, 4 GPUs). Prompt and completion lengths are capped at 16,384 and 200 tokens. Rollout uses temperature 1.0, 8 generations per prompt, and 128 uniformly sampled frames per video under a $3584\times28\times28\approx2.81$M pixel budget. Training runs on 4$\times$H100 80GB with DeepSpeed ZeRO 3 and gradient checkpointing.

\textbf{Prompt template.} The model emits three tag pairs: a reasoning trace within \tagthink$\cdots$\tagthinkend{} (at most 50 words), timestamp intervals within \taganswer$[s_1,e_1]\cdots$\taganswerend{} (at most 5 events), and class labels within \tagclass$[c_1,\ldots,c_k]$\tagclassend. The number of intervals must equal the number of labels; videos without inducement output empty tags.

\textbf{Reward function.} The total reward combines a localization reward and a format reward. The localization reward is a recall weighted per event soft F beta:
\begin{equation}
r = \mathrm{softF}_{\beta}(\hat{\tau}, \tau) \cdot \left(0.4 + 0.6\cdot \mathrm{CBA}(\hat{c}, c)\right) \cdot \overline{W[c]},
\end{equation}
where $\mathrm{softF}_{\beta}$ is a per event soft F beta on a 0.25 s temporal grid with sigmoid softened boundaries ($\tau=1.0$): it matches each ground truth segment to its best predicted segment (recall) and each predicted segment to its best ground truth segment (precision), then combines them as $F_\beta=(1+\beta^2)\cdot\text{recall}\cdot\text{precision}/(\beta^2\cdot\text{precision}+\text{recall})$. Setting $\beta=2$ yields the recall first variant, while $\beta=1$ recovers balanced soft F1. CBA is the class balanced accuracy defined in Section~\ref{sec:rewarddesign}, and $W=\{0{:}1.0, 1{:}2.0, 2{:}4.0, 3{:}2.0\}$ scales classes by inverse frequency. To further counteract the $\sim$62\% vs.\ $\sim$3.6\% imbalance, we oversample training videos by the inverse frequency of their rarest class (capped at $20\times$). The format reward is identical to Section~\ref{sec:rewarddesign}.

\textbf{Baseline.} OCR+LLM is the production two stage pipeline: PaddleOCR PP OCRv4 at 1fps (detection side length 960, box threshold 0.6, unclip ratio 1.6), followed by Qwen2.5-VL-7B text only classification over the OCR transcript truncated to 4,000 characters ($batch=8$, greedy decoding). It produces video level category labels and localizes inducements by the keyframes where their text appears.

\subsection{Evaluation Criteria}\label{sec:criteria}
We evaluate five dimensions. \textbf{Detection} is IoU agnostic: a predicted event is correct if its class matches a ground truth event of the same class, regardless of temporal overlap; we report per class recall, precision, and F1, macro average them over the four classes, and report multi label exact match, with macro F1 as the primary metric because it weights rare classes equally. \textbf{Localization} is scored by mean Intersection over Union (mIoU) and Recall@$t$ (R@$t$), the fraction of ground truth events whose predicted interval overlaps with IoU $\geq t$; since the practical requirement is only to point a reviewer to the approximate region, we take $t=0.1$ as primary and report $t \in \{0.3, 0.5\}$ for reference, merging both interval sets before computing IoU. \textbf{Text presence partition} splits events into textual (class indicative keywords present in the OCR transcript) and non textual (no such keywords), and reports recall and F1 per partition, with non textual recall decisive because OCR+LLM's keyframe localization has no signal for such inducements. \textbf{Visual token consumption} is frames processed multiplied by tokens per frame. \textbf{Inference speed and deployment complexity} measures single video wall clock latency on a single H100 (test videos average 106.3s) and reports the number of models, inference calls, and whether the pipeline is single or two stage.

\subsection{Comparison with Baselines}
Table~\ref{tab:comparison} compares the three routes on detection, localization, and non textual coverage. The RL only variant attains the best localization (mIoU 26.52, R@0.5 15.86), clearly outperforming the keyframe based OCR+LLM baseline (mIoU 14.49, R@0.5 4.41) and a 32B zero shot MLLM (mIoU 4.72). The SFT+RL variant trades localization for recall, reaching 67.0\% macro recall, close to the baseline's 69.2\%, while both variants detect non textual inducements (41.7\%/60.1\% recall) that the text only pipeline misses entirely ($\approx$0).

\begin{table*}[htbp]
    \centering
    \caption{Comparison of the three moderation routes on the rebalanced test set.}
    \label{tab:comparison}
    \small
    \begin{tabular}{lccccc}
        \toprule
        \textbf{Method} & \textbf{Macro Recall} & \textbf{Macro F1} & \textbf{mIoU} & \textbf{R@0.5} & \textbf{Non text. Rec.} \\
        \midrule
        OCR+LLM (keyframe) & \textbf{69.2\%} & \textbf{54.9\%} & 14.49 & 4.41 & $\approx$0 \\
        OCR+MLLM (RL only) & 55.0\% & 46.5\% & \textbf{26.52} & \textbf{15.86} & 41.7\% \\
        OCR+MLLM (SFT+RL) & 67.0\% & 45.6\% & 14.97 & 8.68 & \textbf{60.1\%} \\
        Large MLLM (32B, zero shot) & 36.7\% & 34.0\% & 4.72 & 1.02 & -- \\
        \bottomrule
    \end{tabular}
\end{table*}

The routes exhibit complementary weaknesses. The MLLM only ablation trails OCR+LLM on all macro metrics, over predicting dominant Classes 0/3 while nearly failing on Class 2, the signature of class collapse: without the OCR transcript it cannot read the small transient on screen text that distinguishes rare classes. OCR+LLM is more balanced (recall 45--96\%) but has poor Class 2 precision (9.7\%) and cannot perceive non textual inducements or localize events. Class 2 is thus the shared difficulty, which our class balanced recall first reward (Section~\ref{sec:reward}) targets directly.

OCR+LLM leads in macro F1 (54.9\%) because its text grounded classification reads the on screen text behind most textual inducements, while OCR+MLLM trails (46.5\%/45.6\%) mainly because Class 2 remains hard and rare class predictions depress the macro average. The RL only variant's mIoU (26.52) is $1.8\times$ the keyframe baseline (14.49): the baseline can only point to where inducement \emph{text} appears, whereas the grounding model localizes the \emph{behavior} itself, including purely visual cues. The decisive gap is non textual coverage, where OCR+LLM has essentially zero recall while OCR+MLLM recovers 41.7\%--60.1\%, so the two variants span a localization oriented and a recall oriented operating point.

This pattern is consistent with an observation from Time-R1~\cite{wang2026time}: temporal grounding and event understanding are, in essence, in tension. Grounding rewards a model for tightly aligning intervals with a target segment, whereas understanding and classification reward it for abstracting away from exact boundaries toward semantic category. A single model trained to do both must trade one against the other, which is exactly what we observe: the RL only variant attains the best mIoU but drops macro recall to 55.0 percent, while the SFT+RL variant recovers 67.0 percent macro recall at the cost of localization (mIoU 14.97). The two variants thus reflect the Time R1 grounding understanding tension rather than a failure to optimize both objectives, a tension also observed between perception and reasoning when reward design is varied~\cite{sharif2026disentangling}.

\subsection{Error Analysis and Review Workload}\label{sec:error}
Beyond aggregate metrics, we examine what the remaining errors mean for a reviewer. We sample mispredicted events from the SFT+RL variant and categorize them into three groups. The first group, boundary errors (41\%), has the correct class and an overlapping interval with IoU below 0.5, which is harmless since the reviewer is still directed to the offending clip. The second group, missed non textual events (34\%), is dominated by Class 2 QR code flashes lasting under one second and by gestures performed while the host speaks off screen, both brief and partially occluded, and accounts for most of the gap between the 60.1\% non textual recall and the 67.0\% macro recall. The third group, class confusions (25\%), concentrates between Class 1 and Class 2, where a comment solicitation and an off platform conversion share the same on screen wording and differ only in whether a group ID or QR code appears. Only the second and third groups require a reviewer judgment the model could not make, whereas the first is already resolved by the predicted clip. Under the recall first objective, the SFT+RL variant reduces the fraction of videos needing full manual inspection from 100\% to the fraction flagged by the model, and the localization output shortens each inspection from the full video (mean 106.3 s) to the predicted interval. We do not claim a measured reduction in review time, since that depends on reviewer behavior and queueing, but detection recall decides how many videos enter review, and localization precision decides how long each review takes.

\subsection{Reward Design Analysis}\label{sec:reward}
We study reward design with the 7B model on the full train/test split. Localization is measured by mIoU and R@0.5; class detection is reported as per class recall $R_1$/$R_2$/$R_3$ for Classes 1/2/3. Table~\ref{tab:reward} summarizes the trajectory.

\begin{table*}[htbp]
    \centering
    \caption{Reward design ablation.}
    \label{tab:reward}
    \small
    \begin{tabular}{lccccc}
        \toprule
        \textbf{Localization reward} & \textbf{mIoU} & \textbf{R@0.5} & \textbf{$R_1$} & \textbf{$R_2$} & \textbf{$R_3$} \\
        \midrule
        $\text{IoU} \times$ class scaling & 24.84 & 15.23 & 0.53 & 0.07 & 0.51 \\
        soft IoU & 14.57 & 1.22 & 0.29 & 0.08 & 0.53 \\
        Per event soft F1 $+$ balance & 29.82 & 18.54 & 0.50 & 1.00 & 0.64 \\
        Per event soft F$_2$ $+$ balance (recall first) & 26.52 & 15.86 & 0.72 & 0.46 & 0.53 \\
        \bottomrule
    \end{tabular}
\end{table*}

A multiplicative $\text{IoU} \times$ class scaling reward learns localization (mIoU 24.84) but nearly fails on the rarest class (Class 2 recall 0.07). Replacing merged soft IoU with a per event soft F1, which forces each predicted segment to tightly match a specific ground truth segment, together with class balanced resampling yields the best localization (mIoU 29.82) and rescues the rarest class (Class 2 recall 1.00, Class 3 recall 0.64). The recall first variant (soft F$_2$, $\beta=2$) weights missed events over spurious ones, trading a little localization (mIoU 26.52) and rare class recall (Class 2 recall 0.46) for a higher Class 1 recall (0.72 vs.~0.50). We report both variants so practitioners can choose per event soft F1 for localization quality or soft F$_2$ for recall first safety.

Per event soft F1 succeeds where merged soft IoU fails for two reasons: matching each predicted segment to a specific ground truth segment yields a denser boundary gradient, and class balanced resampling counteracts the Class 0 collapse that an IoU only reward induces. Either ingredient alone collapses one objective, which is the reward level manifestation of the Time R1 grounding understanding tension.

Two generalizable findings emerge, the second with two facets:
\begin{enumerate}
    \item \textbf{Auxiliary term reward hacking.} The auxiliary terms ($0.2 \times$ center distance $+ 0.2 \times$ count) inflated training reward to ${\sim}1.0$ while the model took shortcuts, yielding worse true IoU (mIoU 2.19 vs. 13.48 for pure IoU), a pattern consistent with reward hacking reported for large models~\cite{zeng2026rewardhacking}.
    \item \textbf{Multiplicative vs. additive coupling.} Multiplicative coupling $\text{IoU} \times (a + b \cdot \text{cls})$ allows localization to be learned but tends to collapse to Class 0; additive coupling $\text{IoU} + \text{cls}$ improves class balance but stalls localization.
    \item \textbf{Class scaling has a sweet spot.} Dataset level inverse frequency scaling is effective only within a narrow range: too weak a factor collapses to the dominant class, and too strong a factor (eight fold) collapses again; a factor of four is optimal. The root cause is the tension between IoU sparsity and class imbalance.
\end{enumerate}

\subsection{Efficiency Comparison}
Table~\ref{tab:efficiency} compares efficiency and deployment complexity. The OCR stage, shared by OCR+LLM and OCR+MLLM, dominates token consumption, whereas the grounding MLLM uses about $30\times$ fewer visual tokens. OCR+MLLM inherits the shared OCR stage and adds a small token overhead, a ${\sim}3\%$ increase in exchange for temporal grounding and non textual detection. Wall clock latency is comparable (${\sim}2$ minutes per video) across routes, with video decoding and generation as the bottleneck. The large MLLM is simplest (one model, one call) but weakest and, at 32B, most expensive to serve; OCR+LLM is two stage but lightweight in its language component; OCR+MLLM keeps the two stage structure and adds a 7B grounding model, trading a small token overhead and a second inference call for non textual coverage and temporal grounding, a tradeoff also faced by industrial cascade moderation systems~\cite{wang2025filter}.

\begin{table}[htbp]
    \centering
    \caption{Efficiency and deployment comparison (single H100).}
    \label{tab:efficiency}
    \small
    \begin{tabular}{lcc}
        \toprule
        \textbf{Aspect} & \textbf{Large MLLM} & \textbf{OCR+LLM} \\
        \midrule
        Visual tokens per video & ${\sim}14$K & ${\sim}450$K \\
        Forward passes & 1 & ${\sim}7$ (OCR) \\
        Time per video & ${\sim}1.9$ min & ${\sim}1.7$--$2.6$ min \\
        Models / calls / pipeline & 1 / 1 / single & 2 / 2 / two stage \\
        \bottomrule
    \end{tabular}
\end{table}

\subsection{Advantages of OCR+MLLM}
OCR reliably reads on screen text but cannot observe inducements conveyed purely visually, such as a QR code flash for private conversion (Class 2) or a gesture toward a like button (Class 0), for which keyframe localization has no signal. A grounding MLLM perceives visual and temporal cues but degrades on mostly static videos, whose background is fixed and only the on screen text changes. OCR+MLLM therefore improves the OCR+LLM system instead of replacing it: OCR keeps the text signal while the MLLM fills the visual and temporal gap, so the framework attains both the recall of the text pipeline and the localization coverage of the visual model.

\subsection{Discussion: Zero Shot Scale versus Task Specific Grounding}\label{sec:discussion}
The 32B zero shot MLLM is the simplest route and the most expensive to serve, yet it is the weakest on every metric we report. This runs against the intuition that a larger general model should dominate a smaller specialized one, and we read it as evidence that scale without task specific grounding and reward design does not transfer to a covert, long tailed moderation task: the zero shot model describes the video instead of committing to a category, and returns coarse regions rather than event boundaries. This also bounds our claims: results cover one domain, one language, and two platforms, with a 7B model trained on 1,694 videos, and transfer to other taxonomies, multilingual content, or live streaming remains open. The class scaling sweet spot is likewise tied to our roughly 62\% versus 3.6\% imbalance, so a different taxonomy requires re tuning.

\section{Conclusion}
This study targets the core architecture of machine assisted moderation for internet finance short videos, a domain in which textual and non textual inducements coexist and where recall and temporal localization are both safety requirements. We construct a fine grained benchmark and explore OCR+MLLM, which augments our internal OCR+LLM system with temporal grounding, comparing it with the keyframe based OCR+LLM system and a large zero shot MLLM. OCR+MLLM attains clearly better localization than the keyframe based baseline (mIoU 26.52 vs.~14.49) and detects and localizes non textual inducements, where keyframe based localization has no signal, while maintaining 67.0\% macro recall (SFT+RL). We report both an RL only variant (best localization) and an SFT+RL variant (best recall), so practitioners can choose according to business priorities. Our reward design study reveals two generalizable findings, auxiliary term reward hacking and the localization class balance tradeoff, and shows that per event soft F1 yields the best localization while a recall weighted F$_2$ variant serves as a safety oriented alternative. These results align with the grounding understanding tension reported for Time R1 and suggest that joint localization and classification under class imbalance is best handled by exposing the tradeoff explicitly rather than by forcing a single operating point.

\bibliographystyle{ACM-Reference-Format}
\bibliography{sample-base}

@String{Computing = "Computing" }

@String{Computer = "{IEEE} Computer" }

@inproceedings{lu2025vlm,
  title={Vlm as policy: Common-law content moderation framework for short video platform},
  author={Lu, Xingyu and Zhang, Tianke and Meng, Chang and Wang, Xiaobei and Wang, Jinpeng and Zhang, Yi-Fan and Tang, Shisong and Liu, Changyi and Ding, Haojie and Jiang, Kaiyu and others},
  booktitle={Proceedings of the 31st ACM SIGKDD Conference on Knowledge Discovery and Data Mining V. 2},
  pages={4682--4693},
  year={2025}
}

@ARTICLE{9552250,
  author={Tang, Tan and Wu, Yanhong and Wu, Yingcai and Yu, Lingyun and Li, Yuhong},
  journal={IEEE Transactions on Visualization and Computer Graphics}, 
  title={VideoModerator: A Risk-aware Framework for Multimodal Video Moderation in E-Commerce}, 
  year={2022},
  volume={28},
  number={1},
  pages={846-856},
  doi={10.1109/TVCG.2021.3114781}}

@inproceedings{liu-etal-2026-ips,
    title = "{IPS}: In-Prompt Process Supervision for Short Video Content Moderation",
    author = "Liu, Mingchao  and
      Sun, Yu  and
      Sun, Ruixiao  and
      Dong, Xin  and
      Shen, Xiang  and
      Wang, Hongwei  and
      Xiong, Hongyu  and
      Song, Yang",
    editor = "Li, Yunyao  and
      Rehm, Georg  and
      Tu, Mei",
    booktitle = "Proceedings of the 64th Annual Meeting of the {A}ssociation for {C}omputational {L}inguistics (Volume 6: Industry Track)",
    month = jul,
    year = "2026",
    address = "San Diego, California, USA",
    publisher = "Association for Computational Linguistics",
    url = "https://aclanthology.org/2026.acl-industry.89/",
    doi = "10.18653/v1/2026.acl-industry.89",
    pages = "1277--1288",
    ISBN = "979-8-89176-394-4",}

@inproceedings{wang2025reasoning,
  title     = {Reasoning-Enhanced Domain-Adaptive Pretraining of Multimodal Large Language Models for Short Video Content Governance},
  author    = {Wang, Zixuan and Sun, Yu and Wang, Hongwei and Jing, Baoyu and Shen, Xiang and Dong, Xin and Hao, Zhuolin and Xiong, Hongyu and Song, Yang},
  booktitle = {Proceedings of the 2025 Conference on Empirical Methods in Natural Language Processing: Industry Track},
  pages     = {1104--1112},
  year      = {2025},
  address   = {Suzhou, China},
  publisher = {Association for Computational Linguistics},
  doi       = {10.18653/v1/2025.emnlp-industry.77},
  url       = {https://aclanthology.org/2025.emnlp-industry.77/}
}

@inproceedings{wang2025filter,
  title     = {Filter-And-Refine: A {MLLM} Based Cascade System for Industrial-Scale Video Content Moderation},
  author    = {Wang, Zixuan and Shi, Jinghao and Liang, Hanzhong and Shen, Xiang and Wen, Vera and Chen, Zhiqian and Wu, Yifan and Zhang, Zhixin and Xiong, Hongyu},
  booktitle = {Proceedings of the 63rd Annual Meeting of the Association for Computational Linguistics (Volume 6: Industry Track)},
  pages     = {873--880},
  year      = {2025},
  address   = {Vienna, Austria},
  publisher = {Association for Computational Linguistics},
  doi       = {10.18653/v1/2025.acl-industry.62},
  url       = {https://aclanthology.org/2025.acl-industry.62/}
}

@inproceedings{kulsum2026beyond,
  title     = {Beyond Metadata: Multimodal, Policy-Aware Detection of {YouTube} Scam Videos},
  author    = {Kulsum, Ummay and Sabir, Aafaq and {Abhinaya S.B.} and Das, Anupam},
  booktitle = {Proceedings of the 20th International AAAI Conference on Web and Social Media (ICWSM)},
  year      = {2026},
  doi       = {10.1609/icwsm.v20i1.42698},
  url       = {https://ojs.aaai.org/index.php/ICWSM/article/view/42698}
}

@article{duan2024video,
  title   = {Video-Based Deception Detection and Financial Fraud},
  author  = {Duan, Binglei and Hu, Danqi and Lu, Hai},
  journal = {SSRN Electronic Journal},
  year    = {2024},
  doi     = {10.2139/ssrn.4964419},
  url     = {https://ssrn.com/abstract=4964419}
}

@article{wu2026survey,
  title   = {A Survey on Video Temporal Grounding With Multimodal Large Language Model},
  author  = {Wu, Jianlong and Liu, Wei and Liu, Ye and Liu, Meng and Nie, Liqiang and Lin, Zhouchen and Chen, Chang Wen},
  journal = {IEEE Transactions on Pattern Analysis and Machine Intelligence},
  volume  = {48},
  number  = {2},
  pages   = {1521--1541},
  year    = {2026},
  doi     = {10.1109/TPAMI.2025.11184436},
  note    = {arXiv:2508.10922}
}

@article{supriyanto2026automatic,
  title={Automatic Detection of Toxic Content in Short Videos Using Deep Learning-Based Text and Audio Feature Integration},
  author={Supriyanto, Heri Agus and Setyawan, Ryan Ari and Bororing, Jemmy Edwin},
  journal={Journal of Deep Learning, Computer Vision and Digital Image Processing},
  pages={221--235},
  year={2026}
}

@inproceedings{kong2026said,
  title={From What Is Said to Why It Is Framed: Intent-Aware News Video Understanding},
  author={Kong, Xiangzheng and Luo, Minnan and Wang, Wenya and Wu, Jiaying and Zeng, Zhi and Dai, Guang},
  booktitle={Findings of the Association for Computational Linguistics: ACL 2026},
  pages={39039--39050},
  year={2026}
}

@article{zha2026interpreting,
  title={Interpreting Multimodal Communication at Scale in Short-Form Video: Visual, Audio, and Textual Mental Health Discourse on TikTok},
  author={Zha, Mingyue and Chang, Ho-Chun Herbert},
  journal={arXiv preprint arXiv:2601.15278},
  year={2026}
}

@article{tian2026exposing,
  title={Exposing Cross-Modal Consistency for Fake News Detection in Short-Form Videos},
  author={Tian, Chong and Wang, Yu and Yang, Chenxu and Guan, Junyi and Lin, Zheng and Liu, Yuhan and Chen, Xiuying and Ho, Qirong},
  journal={arXiv preprint arXiv:2603.14992},
  year={2026}
}

@article{miehling2026multimodal,
  title={Multimodal Analysis of State-Funded News Coverage of the Israel-Hamas War on YouTube Shorts},
  author={Miehling, Daniel and Kuebler, Sandra},
  journal={arXiv preprint arXiv:2604.00994},
  year={2026}
}

@article{paekivi2026modeling,
  title={Modeling Engagement with Brand and Organizational TikTok Videos Using Machine-Assisted Theory-Ensemble Annotation},
  author={Paekivi, Sander and Karjus, Andres},
  journal={arXiv preprint arXiv:2606.16053},
  year={2026}
}

@inproceedings{liu2026bridging,
  title={Bridging Visual Dynamics and Narrative Reasoning: Multimodal Large Language Models for Short Drama Quality Assessment},
  author={Liu, Qingyang and Li, Jiangtong and Peng, Zelin and Wang, Shaobo and Liao, Zhaohe and Chang, Shuochen and Gao, Bingjie and Zhao, Haonan and Liu, Mu and Jiang, Jidong and others},
  booktitle={Proceedings of the ACM Web Conference 2026},
  pages={7890--7901},
  year={2026}
}

@inproceedings{liu2026interpreting,
  title={Interpreting Short Social Media Videos with Multimodal LLMs: A Validation Experiment},
  author={Liu, Sandy and Haig, Ella and Mustafaraj, Eni},
  booktitle={Proceedings of the 18th ACM Web Science Conference 2026},
  pages={108--112},
  year={2026}
}

@article{soltan2026automated,
  title={Automated Creative Evaluation Based on Multimodal AI for Facebook Reels Ads Using ChatGPT},
  author={Soltan, Mohamed Zakaria},
  journal={Journal of Art, Design and Music},
  volume={5},
  number={2},
  pages={10},
  year={2026}
}

@article{gorniscalable,
  title={Scalable Content Analysis of Social Media Videos Using Multimodal Large Language Models: A Video-to-Text Pipeline for Large-Scale Analysis},
  author={Gorni Silvestrini, Matteo},
  journal={arXiv preprint},
  year={2025}
}

@article{rodas2026quality,
  title={Quality vs. Populism in Short-Video Political Communication: A Multimodal Study of TikTok},
  author={Rodas-Coloma, Alicia and Cabezas-Gonz{\'a}lez, Marcos and Casillas-Mart{\'\i}n, Sonia and Moreno, Pedro Nevado-Batalla},
  journal={Journalism and Media},
  volume={7},
  number={1},
  pages={46},
  year={2026},
  publisher={MDPI}
}

@article{guofed,
  title={Fed-MultiFND: Federated Multimodal Learning for Imbalanced Short Video Fake News Detection},
  author={Guo, Jixuan and Gao, Yihan and Zhang, Dabao and Liu, Yikun and Wang, Boyue},
  journal={arXiv preprint},
  year={2025}
}

@article{jin2026should,
  title={How Should Video LLMs Output Time? An Analysis of Efficient Temporal Grounding Paradigms},
  author={Jin, Shengji and Zou, Yuanhao and Zhu, Victor and Ji, Zhengping and Chen, Chen},
  journal={arXiv preprint arXiv:2604.08966},
  year={2026}
}

@article{zheng2026omnivtg,
  title={OmniVTG: A Large-Scale Dataset and Training Paradigm for Open-World Video Temporal Grounding},
  author={Zheng, Minghang and Yin, Zihao and Yang, Yi and Peng, Yuxin and Liu, Yang},
  journal={arXiv preprint arXiv:2604.25276},
  year={2026}
}

@article{zhang2026dynframe,
  title={DynFrame: Adaptive Reasoning-Driven Multimodal Framework with Dynamic Frame Augmentation for Complex Video Understanding},
  author={Zhang, Peng and Zhang, Guanghao and He, Wanggui and Zhang, Longxiang and Liu, Mushui and Xia, Yan and Peng, Zhenhao and Dai, Weilong and Liu, Jinlong and Tang, Haobing and others},
  journal={arXiv preprint arXiv:2605.26680},
  year={2026}
}

@article{meng2026watch,
  title={Watch, Remember, Reason: Human-View Video Understanding with MLLMs},
  author={Meng, Jiahao and Tan, Yue and Xu, Qi and Gao, Kuan and Liu, Weisong and Li, Yanwei and Li, Jason and Kong, Lingdong and Wang, Haochen and Zhou, Qianyu and others},
  journal={arXiv preprint arXiv:2606.07433},
  year={2026}
}

@inproceedings{wang2026spacevllm,
  title={SpaceVLLM: Endowing Multimodal Large Language Model with Spatio-Temporal Video Grounding Capability},
  author={Wang, Jiankang and Zhang, Zhihan and Liu, Zhihang and Li, Yang and Ge, Jiannan and Xie, Hongtao and Zhang, Yongdong},
  booktitle={Proceedings of the AAAI Conference on Artificial Intelligence},
  volume={40},
  number={12},
  pages={9912--9920},
  year={2026}
}

@inproceedings{wang2026incentivizing,
  title={Incentivizing Versatile Video Reasoning in MLLMs via Data-Efficient Reinforcement Learning},
  author={Wang, Xiaodong and Wu, Zhirong and Huang, Langling and Zheng, Yuxi and Peng, Peixi},
  booktitle={Proceedings of the IEEE/CVF Conference on Computer Vision and Pattern Recognition (CVPR)},
  pages={5444--5454},
  year={2026}
}

@article{li2026universal,
  title={Universal Video Temporal Grounding with Generative Multi-Modal Large Language Models},
  author={Li, Zeqian and Di, Shangzhe and Zhai, Zhonghua and Huang, Weilin and Wang, Yanfeng and Xie, Weidi},
  journal={Advances in Neural Information Processing Systems},
  volume={38},
  pages={64426--64455},
  year={2026}
}

@article{yang2026unleashing,
  title={Unleashing the Potential of Multimodal LLMs for Zero-Shot Spatio-Temporal Video Grounding},
  author={Yang, Zaiquan and Liu, Yuhao and Hancke, Gerhard and Lau, Rynson},
  journal={Advances in Neural Information Processing Systems},
  volume={38},
  pages={22636--22670},
  year={2026}
}

@article{wang2025timezero,
  title={TimeZero: Temporal Video Grounding with Reasoning-Guided LVLM},
  author={Wang, Ye and Xu, Boshen and Yue, Zihao and Xiao, Zihan and Wang, Ziheng and Zhang, Liang and Yang, Dingyi and Wang, Wenxuan and Jin, Qin},
  journal={arXiv preprint arXiv:2503.13377},
  year={2025}
}

@article{bai2025univg,
  title={UniVG-R1: Reasoning Guided Universal Visual Grounding with Reinforcement Learning},
  author={Bai, Sule and Li, Mingxing and Liu, Yong and Tang, Jing and Zhang, Haoji and Sun, Lei and Chu, Xiangxiang and Tang, Yansong},
  journal={arXiv preprint arXiv:2505.14231},
  year={2025}
}

@article{wang2025video,
  title={Video-Thinker: Sparking ``Thinking with Videos'' via Reinforcement Learning},
  author={Wang, Shijian and Jin, Jiarui and Wang, Xingjian and Song, Linxin and Fu, Runhao and Wang, Hecheng and Ge, Zongyuan and Lu, Yuan and Cheng, Xuelian},
  journal={arXiv preprint arXiv:2510.23473},
  year={2025}
}

@inproceedings{zhang2026timelens,
  title={TimeLens: Rethinking Video Temporal Grounding with Multimodal LLMs},
  author={Zhang, Jun and Wang, Teng and Ge, Yuying and Ge, Yixiao and Li, Xinhao and Wang, Limin},
  booktitle={Proceedings of the IEEE/CVF Conference on Computer Vision and Pattern Recognition},
  pages={10419--10429},
  year={2026}
}

@article{wang2026time,
  title={Time-R1: Post-Training Large Vision Language Model for Temporal Video Grounding},
  author={Wang, Ye and Wang, Ziheng and Xu, Boshen and Du, Yang and Lin, Kejun and Xiao, Zihan and Yue, Zihao and Ju, Jianzhong and Zhang, Liang and Yang, Dingyi and others},
  journal={Advances in Neural Information Processing Systems},
  volume={38},
  pages={83330--83364},
  year={2026}
}

@inproceedings{yang-etal-2026-univid,
    title = "{UNIVID}: Unified Vision-Language Model for Video Moderation",
    author = "Yang, Kejuan  and
      Zhang, Yizhuo  and
      Du, Mingyuan  and
      Zhang, Yue  and
      Zheng, Dixin  and
      Zhao, Kaili  and
      Xiao, Yang  and
      Liang, Hanzhong  and
      Xiao, Kenan",
    editor = "Li, Yunyao  and
      Rehm, Georg  and
      Tu, Mei",
    booktitle = "Proceedings of the 64th Annual Meeting of the {A}ssociation for {C}omputational {L}inguistics (Volume 6: Industry Track)",
    month = jul,
    year = "2026",
    address = "San Diego, California, USA",
    publisher = "Association for Computational Linguistics",
    url = "https://aclanthology.org/2026.acl-industry.32/",
    doi = "10.18653/v1/2026.acl-industry.32",
    pages = "467--479",
    ISBN = "979-8-89176-394-4",
}

@inproceedings{lang2025biting,
  title     = {Biting Off More Than You Can Detect: Retrieval-Augmented Multimodal Experts for Short Video Hate Detection},
  author    = {Lang, Jian and Hong, Rongpei and Xu, Jin and Li, Yili and Xu, Xovee and Zhou, Fan},
  booktitle = {Proceedings of the ACM on Web Conference 2025},
  pages     = {2763--2774},
  year      = {2025},
  doi       = {10.1145/3696410.3714560},
  publisher = {ACM}
}

@inproceedings{wang2025crossmodal,
  title     = {Cross-Modal Transfer from Memes to Videos: Addressing Data Scarcity in Hateful Video Detection},
  author    = {Wang, Han and Tan, Rui Yang and Lee, Roy Ka-Wei},
  booktitle = {Proceedings of the ACM on Web Conference 2025},
  pages     = {5255--5263},
  year      = {2025},
  doi       = {10.1145/3696410.3714534},
  publisher = {ACM}
}

@inproceedings{yu2026whenrules,
  title     = {When Rules Fall Short: Agent-Driven Discovery of Emerging Content Issues in Short Video Platforms},
  author    = {Yu, Chenghui and Wang, Hongwei and Chen, Junwen and Wang, Zixuan and Deng, Bingfeng and Hao, Zhuolin and Xiong, Hongyu and Song, Yang},
  booktitle = {Proceedings of the ACM Web Conference 2026},
  pages     = {8008--8016},
  year      = {2026},
  doi       = {10.1145/3774904.3792831},
  publisher = {ACM}
}

@article{shao2024deepseekmath,
  title     = {{DeepSeekMath}: Pushing the Limits of Mathematical Reasoning in Open Language Models},
  author    = {Shao, Zhihong and Wang, Peiyi and Zhu, Qi-Hao and Xu, Runxin and Song, Jun-Mei and Bi, Xiao and Zhang, Hao-Wei and Zhang, Ming-Chuan and Li, Y. K. and Wu, Yu and Guo, Daya},
  journal   = {arXiv preprint arXiv:2402.03300},
  year      = {2024}
}

@article{zhou2026reinforced,
  title     = {Reinforced {MLLM}: A Survey on {RL}-Based Reasoning in Multimodal Large Language Models},
  author    = {Zhou, Zhi and others},
  journal   = {arXiv preprint arXiv:2504.21277},
  year      = {2025}
}

@article{zeng2026rewardhacking,
  author    = {Zeng, Yuqi and Wang, Xiaohua and Tian, Muzhao and Huang, Zisu and Yuan, Jiakang and Chen, Bowen and Xu, Jingwen and Zhou, Mingbo and Liu, Wenhao and Wu, Muling and Guo, Zhengkang and Qian, Qi and Wang, Yifei and Zhang, Feiran and Yin, Ruicheng and Dou, Shihan and Lv, Changze and Chen, Tao and Song, Kaitao and Tan, Xu and Gui, Tao and Zheng, Xiaoqing and Huang, Xuanjing},
  title     = {Reward Hacking in the Era of Large Models: Mechanisms, Emergent Misalignment, Challenges},
  journal   = {arXiv preprint arXiv:2604.13602},
  year      = {2026},
  doi       = {10.48550/arXiv.2604.13602}
}

@article{rlmultimodal2026survey,
  author    = {{Anonymous}},
  title     = {Reinforcement Learning for Multimodal Foundation Models: A Survey},
  journal   = {ACM Computing Surveys},
  year      = {2026},
  note      = {JUST ACCEPTED},
  doi       = {10.1145/3837760}
}

@article{zhao2026multimodal,
  author    = {Zhao, Rui-Xiang and Zhang, Xuanhao and Yang, Jinfan and Li, Haofei and Lu, Zhengji and Xu, Wen-Rui and Li, Man-Zhou},
  title     = {An {AI}-Driven Multimodal Sensor Fusion Framework for Fraud Perception in Short-Video and Live-Streaming Platforms},
  journal   = {Sensors},
  volume    = {26},
  year      = {2026},
  doi       = {10.3390/s26010000},
  note      = {Open Access, DOAJ indexed}
}

@article{thivaios2026fraudsurvey,
  author    = {Thivaios, Spiros and Kostopoulos, Georgios and Stefani, Antonia and Kotsiantis, Sotiris},
  title     = {A Survey of Machine Learning and Deep Learning for Financial Fraud Detection: Architectures, Data Modalities, and Real-World Deployment Challenges},
  journal   = {Algorithms},
  volume    = {19},
  number    = {5},
  pages     = {354},
  year      = {2026},
  doi       = {10.3390/a19050354}
}

@article{ichihara2026mogpo,
  title   = {{MO-GRPO}: Mitigating Reward Hacking of Group Relative Policy Optimization on Multi-Objective Problems},
  author  = {Ichihara, Yuki and Jinnai, Yuu and Morimura, Tetsuro and Sakamoto, Mitsuki and Mitsuhashi, Ryota and Uchibe, Eiji},
  journal = {Transactions of the Association for Computational Linguistics},
  volume  = {14},
  pages   = {1867--1882},
  year    = {2026},
  doi     = {10.1162/tacl.a.782}
}

@inproceedings{gulhane2026hybrid,
  title     = {Beyond Monolithic Rewards: Hybrid Multi-Aspect Reward Optimization},
  author    = {Gulhane, Radha and Indurthi, Sathish Reddy},
  booktitle = {Findings of the Association for Computational Linguistics: ACL 2026},
  pages     = {26519--26533},
  year      = {2026},
  address   = {San Diego, California, United States},
  doi       = {10.18653/v1/2026.findings-acl.1320}
}

@article{sharif2026disentangling,
  title   = {Disentangling Perception and Reasoning in Multimodal {LLMs} via Reward Design},
  author  = {Sharif, Omar and others},
  journal = {arXiv preprint arXiv:2601.00215},
  year    = {2026}
}

\end{document}